\PassOptionsToPackage{table}{xcolor}
\documentclass[]{fairmeta}

\usepackage{silence}
\usepackage[utf8]{inputenc} 
\usepackage[T1]{fontenc}    
\usepackage{url}            
\usepackage{booktabs}       
\usepackage{amsfonts}       
\usepackage{nicefrac}       
\usepackage{microtype}      
\usepackage{xcolor}
\usepackage{soul}
\definecolor{bluelink}{RGB}{0,113,188}
\definecolor{greenlink}{RGB}{0,188,113}
\definecolor{PineGreen}{rgb}{0.0, 0.47, 0.44}
\definecolor{Gray}{rgb}{0.5,0.5,0.5}
\definecolor{audio_desc}{rgb}{0.82,0.99,0.80}
\definecolor{shot_desc_1}{rgb}{0.98,0.87,0.87}
\definecolor{shot_desc_2}{rgb}{0.98,0.95,0.83}
\definecolor{shot_desc_3}{rgb}{0.92,0.86,0.98}
\usepackage{listings} 
\usepackage{wrapfig}
\usepackage[most]{tcolorbox}

\newtcolorbox{samplebox}[1]{
    breakable, 
    colback=blue!5!white,
    colframe=blue!50!black, 
    fonttitle=\bfseries,
    title=#1
}
\definecolor{citecolor}{HTML}{0071bc}
\hypersetup{
    colorlinks=true,%
    citecolor=citecolor,%
    filecolor=citecolor,%
    linkcolor=citecolor,%
    urlcolor=citecolor
}
\usepackage{tabularx}
\usepackage{amsmath}
\usepackage{multirow}
\usepackage{array}
\usepackage{enumitem}
\usepackage{tikz}
\usepackage{lipsum}
\usepackage{adjustbox}
\usepackage{siunitx}
\usepackage{graphicx}     
\usepackage{float}

\usepackage{amssymb}
\renewcommand{\paragraph}[1]{\vspace{1.25mm}\noindent\textbf{#1}}

\usepackage{colortbl}       
\definecolor{oursrow}{RGB}{220, 235, 255}

\usepackage{algorithm}      
\usepackage{algorithmic}    

\newcommand{\tablefontsize}{\footnotesize}

\definecolor{claimred}{RGB}{196, 48, 48}      
\definecolor{claimblue}{RGB}{0, 102, 179}     

\definecolor{ao}{rgb}{0.0, 0.0, 1.0}
\definecolor{airforceblue}{rgb}{0.36, 0.54, 0.66}
\definecolor{ceruleanblue}{rgb}{0.16, 0.32, 0.75}
\definecolor{cerulean}{rgb}{0.0, 0.48, 0.65}
\definecolor{celestialblue}{rgb}{0.29, 0.59, 0.82}
\definecolor{azure(colorwheel)}{rgb}{0.0, 0.5, 1.0}
\definecolor{babyblue}{rgb}{0.54, 0.81, 0.94}
\definecolor{babyblueeyes}{rgb}{0.63, 0.79, 0.95}
\definecolor{ballblue}{rgb}{0.13, 0.67, 0.8}

\definecolor{asparagus}{rgb}{0.53, 0.66, 0.42}
\definecolor{ao(english)}{rgb}{0.0, 0.5, 0.0}
\definecolor{applegreen}{rgb}{0.55, 0.71, 0.0}
\definecolor{armygreen}{rgb}{0.29, 0.33, 0.13}
\definecolor{gray-asparagus}{rgb}{0.27, 0.35, 0.27}
\definecolor{green(ryb)}{rgb}{0.4, 0.69, 0.2}

\definecolor{amethyst}{rgb}{0.6, 0.4, 0.8}
\definecolor{antiquefuchsia}{rgb}{0.57, 0.36, 0.51}
\definecolor{blue-violet}{rgb}{0.54, 0.17, 0.89}
\definecolor{brightlavender}{rgb}{0.75, 0.58, 0.89}
\definecolor{brightube}{rgb}{0.82, 0.62, 0.91}
\definecolor{brilliantlavender}{rgb}{0.96, 0.73, 1.0}

\definecolor{amber}{rgb}{1.0, 0.75, 0.0}
\definecolor{amber(sae/ece)}{rgb}{1.0, 0.49, 0.0}
\definecolor{atomictangerine}{rgb}{1.0, 0.6, 0.4}
\definecolor{burntorange}{rgb}{0.8, 0.33, 0.0}
\definecolor{burntsienna}{rgb}{0.91, 0.45, 0.32}
\definecolor{cadmiumorange}{rgb}{0.93, 0.53, 0.18}
\definecolor{carrotorange}{rgb}{0.93, 0.57, 0.13}
\definecolor{chocolate(web)}{rgb}{0.82, 0.41, 0.12}
\definecolor{chromeyellow}{rgb}{1.0, 0.65, 0.0}
\definecolor{darkorange}{rgb}{1.0, 0.55, 0.0}
\definecolor{darktangerine}{rgb}{1.0, 0.66, 0.07}
\definecolor{deepcarrotorange}{rgb}{0.91, 0.41, 0.17}
\definecolor{deepsaffron}{rgb}{1.0, 0.6, 0.2}
\definecolor{fulvous}{rgb}{0.86, 0.52, 0.0}

\newlength\savewidth
\newcolumntype{x}[1]{>{\centering\arraybackslash}p{#1pt}}
\newcolumntype{y}[1]{>{\raggedright\arraybackslash}p{#1pt}}
\newcolumntype{z}[1]{>{\raggedleft\arraybackslash}p{#1pt}}

\usepackage{pgf}
\usepackage{rotating}
\usepackage[abs]{overpic}
\usepackage{makecell}
\usepackage{longtable}
\usepackage{csquotes}
\definecolor{eventcolor}{RGB}{70,130,180}   
\definecolor{shotcolor}{RGB}{218,165,32}    
\definecolor{entitycolor}{RGB}{220,20,60}   
\definecolor{globalcolor}{RGB}{150,210,120}   

\title{SpongeBob: Sync-Aware Harmonious Audio-Visual Generative Editing}
\newcommand{\papertitle}{SpongeBob: Sync-Aware Harmonious Audio-Visual Generative Editing}

\author[1,2\dag]{Sen Liang}
\author[2\dag]{Cong Wang}
\author[1]{Fengbin Guan}
\author[2]{Zhentao Yu}
\author[1]{Yiting Lu}
\author[2]{Yuanzhi Wang}
\author[2]{\\Yuan Zhou}
\author[1*]{Xin Li}
\author[1*]{Zhibo Chen}
\affiliation[1]{University of Science and Technology of China}
\affiliation[2]{Tencent Hunyuan}

\usepackage{fancyhdr}
\fancypagestyle{titlepagewithlogo}{
  \fancyhf{} 
  \rhead{\includegraphics[height=0.6cm]{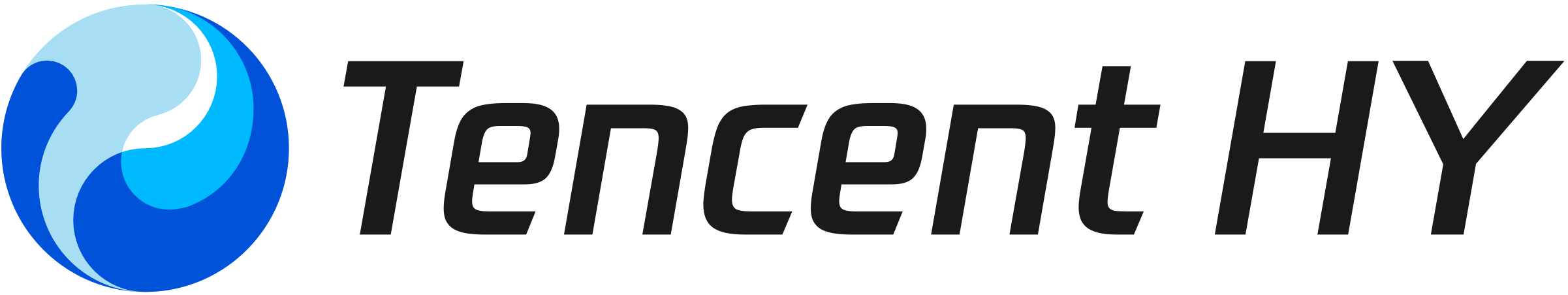}} 
}

\abstract{} 

\begin{document}

\renewcommand{\thefootnote}{\fnsymbol{footnote}}
\thispagestyle{titlepagewithlogo}
\maketitle
\footnotetext[2]{Equal contribution.}
\footnotetext[1]{Corresponding author. Email: chenzhibo@ustc.edu.cn}
\renewcommand{\thefootnote}{\arabic{footnote}}
\vspace{-15mm}
\begin{center}
  \includegraphics[width=\linewidth]{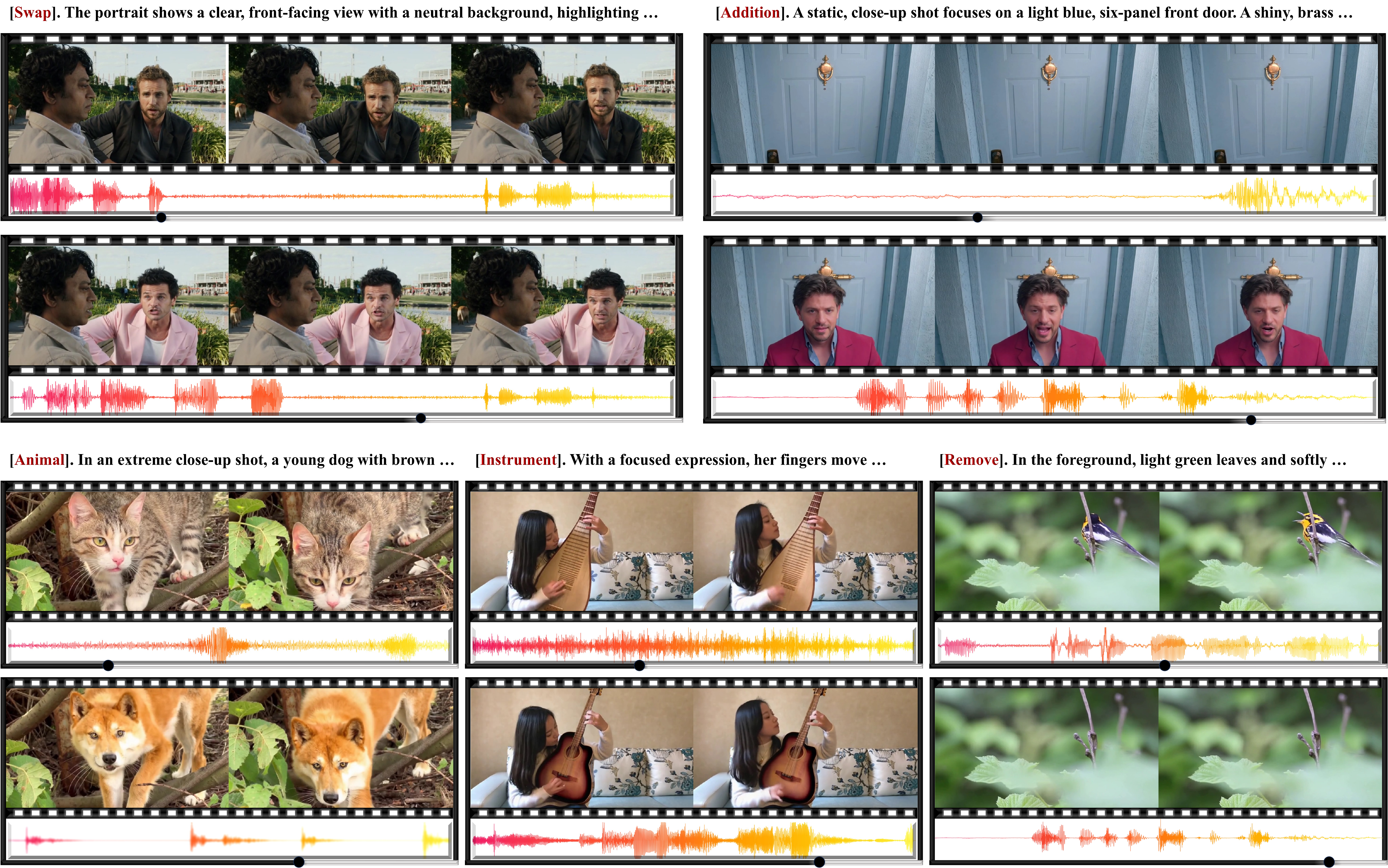}
  \captionsetup{hypcap=false}
  \vspace{-5mm}
  \captionof{figure}{
    \textbf{SpongeBob unifies visual editing and audio synthesis in a unified pass, achieving frame-level synchronization while seamlessly preserving unedited context.}
}
  \label{fig:teaser}
  
\end{center}

\begin{center}
{\large\textbf{Abstract}}
\end{center}
Visual and acoustic events in the physical world are inherently coupled, yet existing video editing methods typically adopt decoupled pipelines, lacking bidirectional modality interaction. This results in two key limitations: (i) audio-visual desynchronization and (ii) contextual conflicts between generated audio and preserved content.
To address these, we propose \textbf{SpongeBob}, the first end-to-end audio-visual joint editing framework featuring bidirectional cross-modal interaction. For synchronization, a Sync-Aware Mechanism aligns visual edits with sound events via bidirectional attention, temporal alignment, and spatial constraints. For contextual consistency, a Context-Aware Module leverages acoustic and visual context attention to prevent semantic clashes. Additionally, we introduce Sync-Preserving Training and Guidance (SPTG) to enhance alignment without degrading quality.
Due to the scarcity of paired data, we construct a scalable data pipeline and a large-scale subject-level dataset. We also propose SpongeBob-Bench for systematic evaluation. Experiments show SpongeBob significantly outperforms existing baselines, improving Sync-C by 30\% and Ctx-F1 by 12.5\%. Our project page is available at \href{https://hy-spongebob.github.io/}{https://hy-spongebob.github.io/}.

\section{Introduction}

Recent advances in diffusion models have revolutionized video editing~\citep{jiang2025vace,liang2025omniv2v,qin2024instructvid2vid,yang2025unified,editverse,ICVE,zi2025se}, enabling precise subject-level manipulation of visual content with high temporal consistency. 
However, in the physical world, visual events are inextricably coupled with their acoustic counterparts; any modification to a visual subject (e.g., deleting a speaking person or exchanging a barking dog for a meowing cat) must be naturally reflected in its synchronized audio stream to maintain physical plausibility. 
Despite the maturity of visual-only editing, without cross-modal synergy, even pixel-level visual fidelity fails to preserve realism. 
Consequently, developing a framework for synchronized audio-visual editing has emerged as a critical demand for next-generation multimodal content creation.

Existing audio-visual editing methods usually fall into two paradigms, both of which separate the two modalities. One line~\citep{jiang2025vace,shan2025hunyuanvideo,ishii2025coherent} uses cascaded pipelines that first complete the visual edit and then generate or repair the audio with other expert models. Another line adopts training-free strategies~\cite{lin2026zero} via cross-modal noise inversion or unidirectional condition injection. Both paradigms can achieve coarse alignment when the edit is small or the audio is generated from scratch, but they share a structural limitation: audio and video are produced in disjoint stages with no feedback loop between them during denoising, causing two characteristic failures. (1) Audio-visual desynchronization: edited lip motion and sound events drift apart at the frame level, speech starts several frames after the mouth opens, or a door-slam sound arrives after the door closes. (2) Audio-visual context conflict: newly generated sounds ignore the unedited audio-visual context, in a two-person dialogue where only speaker A is edited, the regenerated voice may overlap speaker B's unedited turn or break the original turn-taking structure. 
%
Both failures trace to \emph{the lack of an end-to-end framework that supports bidirectional cross-modal interaction during the editing denoising process for closing the perceptual gap.}

However, realizing this end-to-end framework raises two fundamental challenges. 
(1) Data challenge. Conventionally, training an end-to-end editor would require (pre-edit audio-visual content, post-edit audio-visual content, editing instruction) triplets as supervision. 
However, such triplets do not occur naturally at scale, since no web-scale corpus contains the same scene edited in two different ways, while collecting them by hand is prohibitively expensive.
This has historically blocked end-to-end training for this task.
%
%
(2) Architecture challenge. 
Even with data in place, the framework must satisfy two closely related aspects during denoising:
(i) Synchronization modeling: the model must continuously maintain temporal correspondence between target visual motion and sound events during denoising, rather than passively aligning them post-generation; cross-modal interaction must also incorporate spatial constraints so that audio-driven visual changes act only on the target subject region without spreading to the background or other instances. (ii) Context preservation: audio editing must preserve the source context, including background sounds, ambient audio, and non-target speakers, so that newly generated subject-specific sounds coexist harmoniously with the original audio-visual scene, rather than rebuilding the entire audio track from scratch.

In this paper, we present \textbf{SpongeBob}, \emph{a dual-stream Diffusion Transformer (DiT) that addresses both challenges within a single unified framework}. At its core, SpongeBob reformulates audio-visual editing as a self-supervised inpainting task: given any ordinary audio-visual clip, we mask the target subject in both modalities and train the model to reconstruct the original signal conditioned on a textual description of what was masked; at inference, the user provides a different textual description for the masked region, so reconstruction becomes targeted editing. This reformulation replaces the impractical need for (pre-edit, post-edit, instruction) triplets with (clip, mask, caption) examples that can be produced from ordinary single-take videos via automated segmentation, audio separation, and multi-stage filtering, thereby resolving the data challenge and unlocking end-to-end training. Under this formulation, SpongeBob addresses the architecture challenge through three tightly coupled components. Sync-Aware Editing Mechanism targets synchronization modeling, aligning target visual motion and sound events during denoising via bidirectional cross-modal attention (interaction), three-way temporal RoPE unification (temporal correspondence), and mask-guided asymmetric routing (spatial constraints). Context-Aware Module targets context preservation by adding two zero-initialized cross-attention layers, Acoustic Context Attention over the base audio track and Visual Context Attention over the unedited video region, so the generated audio perceives what must be preserved rather than resynthesizing from scratch. Sync-Preserving Training and Guidance (SPTG) activates these capabilities through a multi-task alignment training schedule and a two-stage inference guidance scheme. To our knowledge, SpongeBob is the first framework to integrate bidirectional cross-modal attention within a unified denoising step for subject-level audio-visual editing, in contrast to concurrent cascaded methods that orchestrate separately trained audio and video modules at system level.
%


Our main contributions are summarized as follows:
\begin{enumerate}
    \item Problem reformulation. We \textbf{recast subject-level audio-visual editing from a supervised task into a self-supervised inpainting task} that needs only ordinary audio-visual clips paired with textual descriptions of their content. This reformulation unlocks end-to-end training for a task that has historically been blocked by data scarcity.
    \item Architecture. We propose \textbf{SpongeBob}, the first end-to-end audio-visual joint editing framework based on bidirectional cross-modal interaction, with three key designs:
    Sync-Aware Editing Mechanism addresses desynchronization from interaction, temporal, and spatial dimensions;
    Context-Aware Module addresses context conflict from audio and visual dimensions;
    SPTG protects cross-modal synchronization and context consistency at both training and inference stages.
    \item Data engineering. 
    We build a scalable data pipeline that produces the first large-scale dataset from unlabeled web video for effective training and benchmarking on subject-level audio-visual editing.
\end{enumerate}

\vspace{-1em}
\section{Related Work}
\vspace{-0.5em}

\noindent\textbf{Video Editing.}
Diffusion-based video editing has developed along two main lines.
Mask-guided methods~\citep{jiang2025vace,liang2025omniv2v} achieve precise region-level editing through explicit spatial conditions; they are technically mature but rely on user-provided masks, limiting flexibility.
Instruction-based methods~\citep{qin2024instructvid2vid,yang2025unified,editverse,ICVE, zi2025se, ditto, OpenVE,Insvie-1m, ku2024anyv2v, liu2025stablev2v} infer editing intent directly from text instructions without additional spatial annotations, offering broader applicability.
Although both lines have achieved significant progress in visual editing quality, their task definition remains confined to the visual modality: when the edited object is itself a sound source, the correspondence between visual content and the original audio is broken, with no mechanism to determine how the corresponding sound should change in synchrony.




\noindent\textbf{Audio-Visual Editing.}
Existing audio-visual editing methods fall into three categories.
Zero-shot methods~\citep{lin2026zero} suffer from low frame rates and lack instance-level control.
Cascaded methods~\citep{jiang2025vace,shan2025hunyuanvideo,ishii2025coherent} first edit video then generate or edit audio, where the video editing stage cannot perceive audio.
AVI-Edit~\citep{zheng2025audio} drives video editing with an audio agent, but audio remains a unidirectional condition.
All the above paradigms decouple the two modalities; the fundamental difference of SpongeBob lies in \emph{bidirectional cross-modal interaction within a unified diffusion process}: video motion, target sound, and acoustic context continuously exchange information during denoising, jointly constraining the editing result.

\vspace{-1em}
\section{The SpongeBob Framework}
\vspace{-0.5em}

\begin{figure}[!t]
    \centering
    \includegraphics[width=\linewidth]{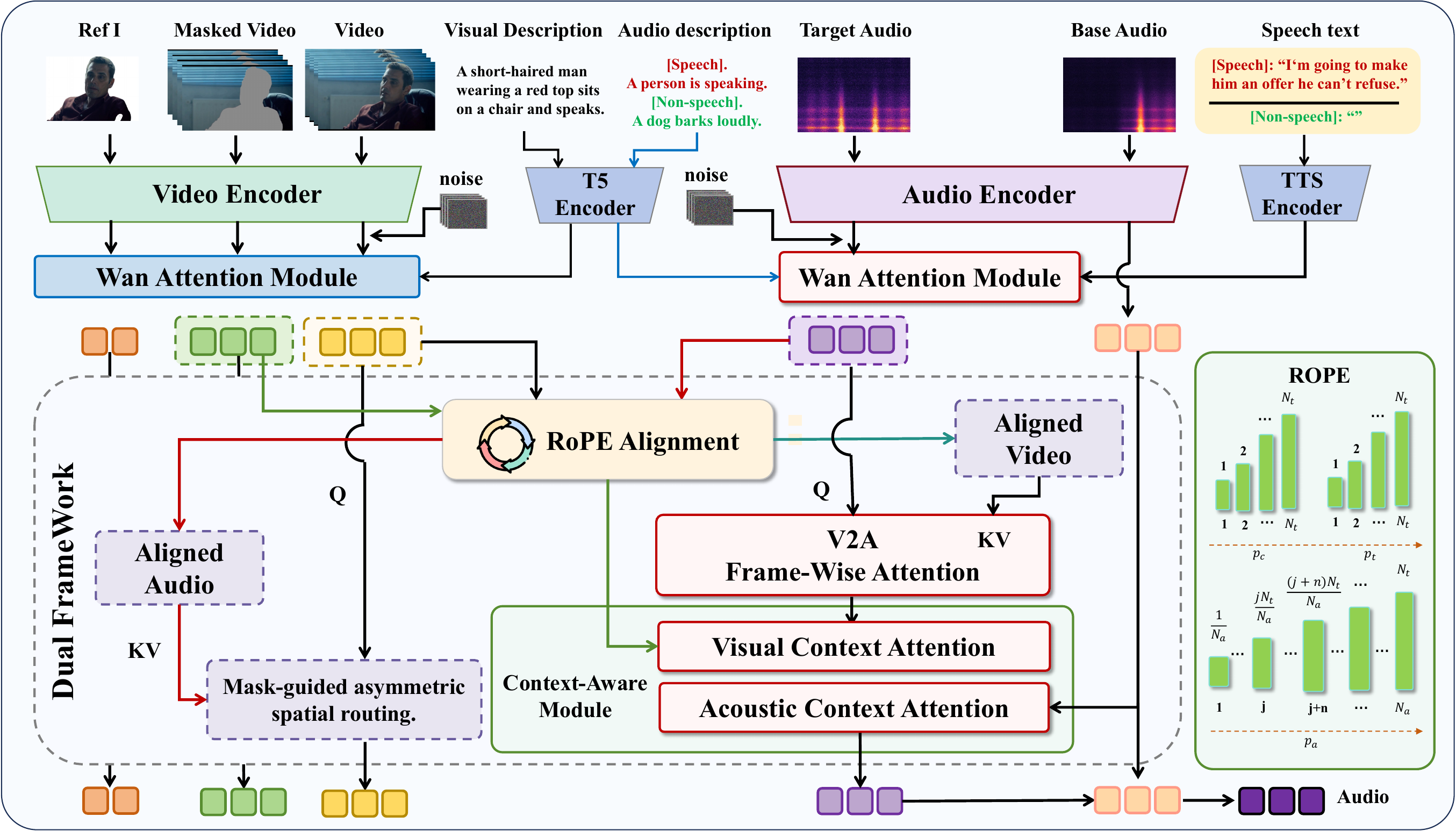}
    \caption{\textbf{Overview of SpongeBob.} Given a source video with an object mask, text instructions, and a reference image, SpongeBob jointly edits the visual content and synthesizes synchronized audio through a dual-stream DiT with sync-aware editing mechanism and context-aware module.}
    \vspace{-1.2em}
    \label{fig:architecture}
\end{figure}

As shown in Fig.~\ref{fig:architecture}, SpongeBob employs a dual-stream Diffusion Transformer (DiT) architecture based on Wan2.2-TI2V-5B~\citep{wan2025wan} that simultaneously edits video and audio within a unified denoising process.
Specifically, for the video stream, the DiT takes a composite latent of a reference image, the masked video (the context), and visual noise as input, while the visual description is injected via cross-attention to guide the reconstruction of the original video clip.
For the audio stream, the model reconstructs the target audio (i.e., the isolated sound of the target subject) from audio noise.
This process is conditioned on the audio description, speech text, and the base audio (i.e., the ambient audio remaining after target audio separation) via dedicated cross-attention layers. 
To ensure effective audio editing, we categorize the target audio into speech and non-speech streams: for speech, the audio description is fixed to a generic prompt (e.g., ``\textit{a person is speaking}'') while the speech text contains the specific linguistic content; for non-speech events, the audio description provides a semantic depiction of the sound (e.g., ``\textit{a dog is barking}'') while the speech text remains null. 
The predicted target audio is finally combined with the base audio to recover the original acoustic signal.
The remainder of this section is organized as follows: \cref{sec:sync_mech} details the Synchronization-Aware Dual-Stream Editing Architecture and its spatial-temporal alignment mechanisms; \cref{sec:context_module} describes the Context-Aware Module for maintaining the consistency of the ambient audio and visual background; \cref{sec:sptg} introduces the Sync-Preserving Training and Guidance (SPTG) strategy for enhanced editing quality; and \cref{sec:data_pipeline} presents our scalable data pipeline.

\vspace{-0.5em}
\subsection{Sync-Aware Editing Mechanism}
\label{sec:sync_mech}
\vspace{-0.5em}

Dual-stream joint denoising provides the foundation for cross-modal interaction, but achieving frame-level audio-visual alignment still requires explicit synchronization mechanisms.
SpongeBob establishes alignment along two dimensions via the sync-aware editing mechanism: \textbf{temporal correspondence} that ensures frame-level synchronization between lip motion, action onset, and sound events; and \textbf{spatial awareness} that confines audio-driven visual changes to the target editing region without disturbing the background or other instances.


\noindent\textbf{RoPE alignment for temporal correspondence.}
A fundamental challenge in synchronized audio-visual editing is the lack of explicit temporal correspondence between heterogeneous token streams.
A sequential assignment of Rotary Positional Embeddings (RoPE) would assign distinct indices to different modalities, treating them as logically separate events even if they occur simultaneously.
To resolve this and enforce cross-modal temporal equivalence, we propose a three-way temporal alignment strategy:
\begin{equation}
    \underbrace{
      p_\text{ref} {=} 0,\;
      p_\text{c}^{(i)} {=} i,\;
      p_\text{t}^{(i)} {=} F{+}i,\;
      p_\text{a}^{(j)} {=} j
    }_{\text{Na\"ive}}
    \;\longrightarrow\;
    \underbrace{
      p_\text{ref} {=} 0,\;
      p_\text{c}^{(i)} {=} i,\;
      p_\text{t}^{(i)} {=} i,\;
      p_\text{a}^{(j)} {=} j {\cdot} \tfrac{N_t}{N_a}
    }_{\text{Ours}}.
    \label{eq:rope}
\end{equation}
where $p_\text{ref}, p_\text{c}, p_\text{t}, p_\text{a}$ denote the temporal indices for the reference image, condition video, target video, and audio streams, respectively.
Specifically, we set $p_\text{ref}=0$ to anchor the static reference image outside the dynamic timeline while maintaining its global accessibility.
The condition and target video tokens share identical temporal indices~$[1, N_t]$ to ensure the temporal alignment, and are distinguished by different diffusion timesteps ($t_\text{cond}{=}0$ vs.\ $t_\text{target}{=}t$).
Most crucially, to align the audio with the video despite the difference in token counts ($N_a \neq N_t$), we map each audio token $j$ to a continuous virtual position $p_\text{a}^{(j)} = j \cdot (N_t/N_a)$, achieving sub-frame temporal synchronization.

\noindent\textbf{Mask-Guided Asymmetric Spatial Routing.}
To achieve precise subject-level control while preventing cross-modal contamination, we implement an asymmetric spatial routing mechanism based on the visual mask $\mathbf{M}$.
Specifically, in the audio-to-video direction, acoustic features are injected strictly into visual tokens within $\mathbf{M}$. 
This localized routing acts as a spatial gate, ensuring that audio-driven updates are confined to the target subject and do not leak into the immutable background.
Conversely, the video-to-audio direction maintains a global receptive field.
This asymmetry is essential because, while the sound source is localized, the acoustic signature is intrinsically shaped by the global context.

\begin{figure*}[!t]
    \centering
    \includegraphics[width=\linewidth]{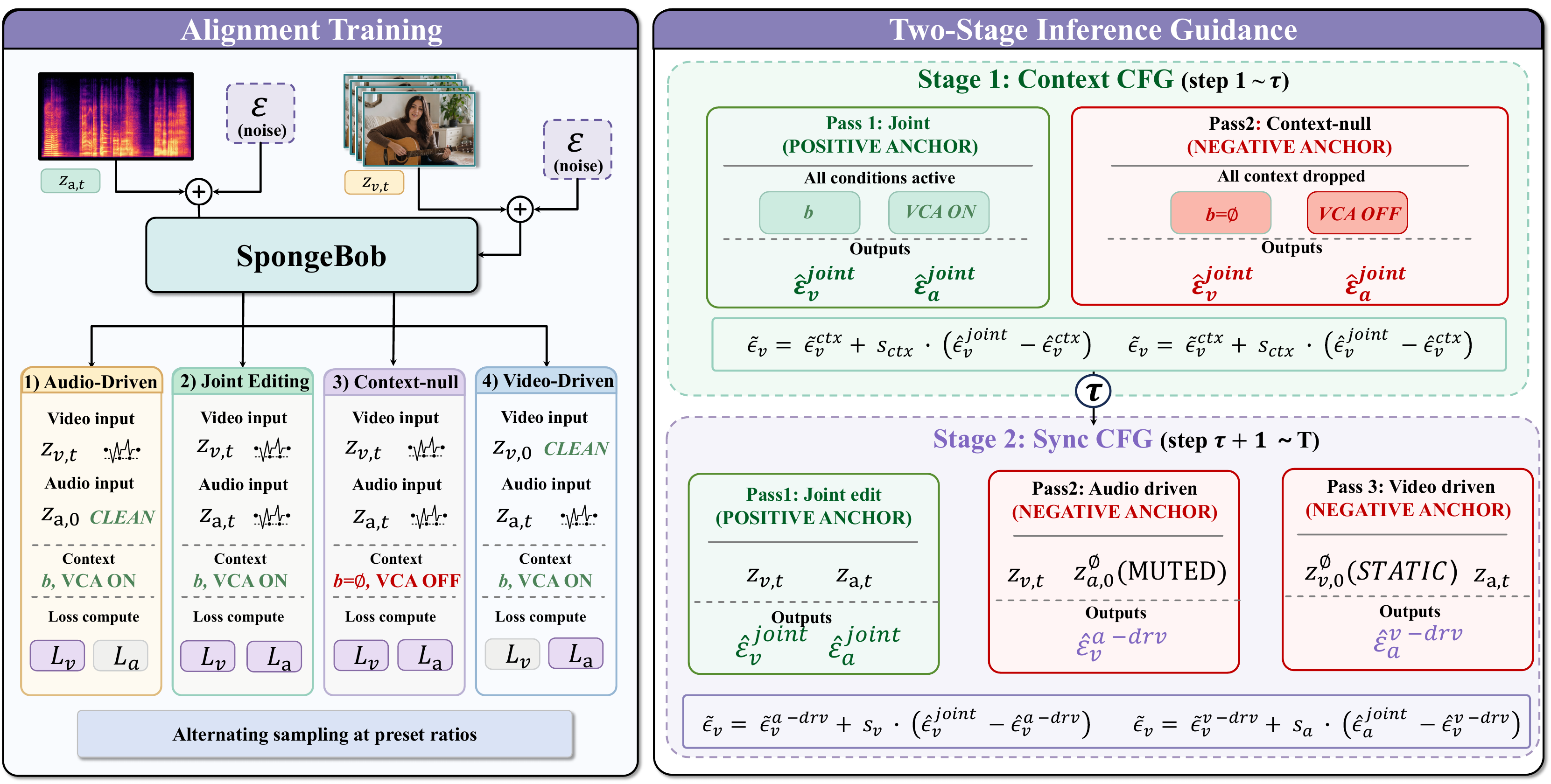}
    \caption{\textbf{Overview of SPTG.} \textbf{Left}: Multi-task alignment training co-trains four modes (joint editing, audio-driven, video-driven, and context-null) to teach stable cross-modal alignment under varied conditioning. \textbf{Right}: Two-stage inference guidance first resolves context conflicts via context CFG (Stage~1), then enhances temporal synchronization via sync CFG (Stage~2).}
    \vspace{-1.5em}
    \label{fig:sptg}
\end{figure*}

\subsection{Context-Aware Module}
\label{sec:context_module}
\vspace{-0.5em}
The synchronization-aware attention in \cref{sec:sync_mech} ensures alignment between modalities but lacks awareness of the unedited environment. 
Without such context, generated subject-specific sounds may conflict with the background (e.g., overlapping with non-target speakers or completely losing the ambient audio). 
To enforce harmonious coexistence, we introduce the Context-Aware Module, which integrates the target audio with the unedited visual and acoustic surroundings through two zero-initialized cross-attention layers.
%

%
First, to ensure the synthesized audio is physically consistent with the visual scene, we enable the target audio stream to perceive the unedited visual background. Formally, we define the Visual Context Attention as:
{
\setlength{\abovedisplayskip}{3pt}
\setlength{\belowdisplayskip}{3pt}
\begin{equation}
    \mathbf{h}_\text{vis} = \text{CrossAttn}(\mathbf{Q} = \mathbf{x}_a^\text{target},\; \mathbf{K} = \mathbf{W}_k^v \mathbf{z}_\text{cond},\; \mathbf{V} = \mathbf{W}_v^v \mathbf{z}_\text{cond}),
\end{equation}
}
where $\mathbf{z}_\text{cond}$ is the masked video feature.
The query originates from the target audio latent, and keys/values are derived from the masked video, enabling audio generation to perceive the visual context beyond the localized editing region.

Second, while visual grounding provides spatial context, avoiding conflicts with existing sounds requires direct perception of the base audio.
Therefore, we formulate the Acoustic Context Attention as:
{
\setlength{\abovedisplayskip}{3pt}
\setlength{\belowdisplayskip}{3pt}
\begin{equation}
    \mathbf{h}_\text{base} = \text{CrossAttn}(\mathbf{Q} = \mathbf{x}_a^\text{target},\; \mathbf{K} = \mathbf{W}_k^b \mathbf{b},\; \mathbf{V} = \mathbf{W}_v^b \mathbf{b}),
    \label{eq:acoustic_ctx}
\end{equation}
}
where $\mathbf{b} \in \mathbb{R}^{N_a \times D_\text{audio}}$ is the base audio encoded by the Audio VAE after source separation.
By attending to $\mathbf{b}$, the target audio stream can perceive the presence of non-target speakers and ambient noise in real-time, allowing it to adapt its energy and timing to avoid overlaps.

\vspace{-1em}
\subsection{Sync-Preserving Training and Guidance (SPTG)}
\label{sec:sptg}
\vspace{-0.5em}

The architecture in the previous sections integrates cross-modal synchronization and context awareness into the denoising process. However, relying solely on joint denoising yields suboptimal results: cross-modal attention falters in high-noise regimes, and standard inference-time CFG enhances text fidelity without explicitly enforcing synchronization or context consistency. As illustrated in Fig.~\ref{fig:sptg}, SPTG addresses these limitations through targeted training and inference strategies.

\vspace{-0.5em}
\subsubsection{Multi-Task Alignment Training}
\label{sec:multi_task}
\vspace{-0.5em}

To mitigate correspondence drift, SpongeBob incorporates three auxiliary objectives alongside the primary editing task. During training, each sample is dynamically routed to one of four modes based on preset probabilities, with the loss computed exclusively for the assigned mode.

\noindent\textbf{Joint Editing.}
Both video and audio targets are denoised at timestep~$t$ with all conditions and context:
{
\setlength{\abovedisplayskip}{3pt}
\setlength{\belowdisplayskip}{3pt}
\begin{equation}
    \mathcal{L}_\text{joint} = \|\boldsymbol{\epsilon}_v - \hat{\boldsymbol{\epsilon}}_v(\mathbf{z}_{v,t}, \mathbf{z}_{a,t}, \mathbf{c}, \mathbf{b})\|^2 + \|\boldsymbol{\epsilon}_a - \hat{\boldsymbol{\epsilon}}_a(\mathbf{z}_{a,t}, \mathbf{z}_{v,t}, \mathbf{c}, \mathbf{b})\|^2.
\end{equation}
}

\noindent\textbf{Audio-driven.}
Audio timestep is set to~$t_a{=}0$ (clean audio), video denoises at~$t_v{=}t$.
The clean audio provides a deterministic anchor for cross-modal attention, mitigating alignment drift.
Only the video loss is computed:
{
\setlength{\abovedisplayskip}{3pt}
\setlength{\belowdisplayskip}{3pt}
\begin{equation}
    \mathcal{L}_\text{a-drv} = \|\boldsymbol{\epsilon}_v - \hat{\boldsymbol{\epsilon}}_v(\mathbf{z}_{v,t}, \mathbf{z}_{a,0}, \mathbf{c}, \mathbf{b})\|^2.
\end{equation}
}

\noindent\textbf{Video-driven.}
Symmetrically, video timestep is set to~$t_v{=}0$, audio denoises at~$t_a{=}t$:
{
\setlength{\abovedisplayskip}{3pt}
\setlength{\belowdisplayskip}{3pt}
\begin{equation}
    \mathcal{L}_\text{v-drv} = \|\boldsymbol{\epsilon}_a - \hat{\boldsymbol{\epsilon}}_a(\mathbf{z}_{a,t}, \mathbf{z}_{v,0}, \mathbf{c}, \mathbf{b})\|^2.
\end{equation}
}

\noindent\textbf{Context-null.}
Both modalities denoise normally with cross-modal attention active, but base audio is nulled ($\mathbf{b} {\to} \varnothing$) and Visual Context Attention is skipped.
The model learns a baseline prediction without context awareness:
{
\setlength{\abovedisplayskip}{3pt}
\setlength{\belowdisplayskip}{3pt}
\begin{equation}
    \mathcal{L}_\text{ctx} = \|\boldsymbol{\epsilon}_v - \hat{\boldsymbol{\epsilon}}_v(\mathbf{z}_{v,t}, \mathbf{z}_{a,t}, \mathbf{c}, \varnothing)\|^2 + \|\boldsymbol{\epsilon}_a - \hat{\boldsymbol{\epsilon}}_a(\mathbf{z}_{a,t}, \mathbf{z}_{v,t}, \mathbf{c}, \varnothing)\|^2.
\end{equation}
}

The sampling probabilities for the four modes are $p_\text{joint}$, $p_\text{a-drv}$, $p_\text{v-drv}$, and $p_\text{ctx}$ respectively. Text conditions and base audio are independently dropped at preset probabilities.

\vspace{-0.5em}
\subsubsection{Two-Stage Inference Guidance}
\label{sec:two_stage}
\vspace{-0.5em}

Unlike standard CFG, which prioritizes textual fidelity without explicitly strengthening context awareness or synchronization, SpongeBob leverages its four trained modes to devise a two-stage guidance strategy. This scheme sequentially addresses context conflicts before enhancing cross-modal alignment.

\noindent\textbf{Stage~1: Context conflict resolution (steps $1{\sim}\tau$).}
Full-conditional and context-null predictions construct a context CFG.
Let $\hat{\boldsymbol{\epsilon}}^{\text{joint}}$ denote the full-conditional prediction and $\hat{\boldsymbol{\epsilon}}^{\text{ctx}}$ the context-null prediction ($\mathbf{b}{\to}\varnothing$, Visual Context Attention skipped):
\begin{equation}
    \tilde{\boldsymbol{\epsilon}}_v = \hat{\boldsymbol{\epsilon}}_v^{\text{ctx}} + s_\text{ctx} \cdot (\hat{\boldsymbol{\epsilon}}_v^{\text{joint}} - \hat{\boldsymbol{\epsilon}}_v^{\text{ctx}}), \qquad
    \tilde{\boldsymbol{\epsilon}}_a = \hat{\boldsymbol{\epsilon}}_a^{\text{ctx}} + s_\text{ctx} \cdot (\hat{\boldsymbol{\epsilon}}_a^{\text{joint}} - \hat{\boldsymbol{\epsilon}}_a^{\text{ctx}}).
\end{equation}
The guidance direction isolates the contribution of the Context-Aware Module, ensuring the generated audio respects unedited content and avoids context conflicts.
This stage requires only 2 forward passes.

\noindent\textbf{Stage~2: Temporal synchronization enhancement (steps $\tau{+}1{\sim}T$).}
Since clean target audio/video is unavailable at inference, muted audio~$\mathbf{z}_{a,0}^{\varnothing}$ and static video~$\mathbf{z}_{v,0}^{\varnothing}$ serve as negative anchors via the audio-driven and video-driven pathways:
\begin{align}
    \tilde{\boldsymbol{\epsilon}}_v &= \hat{\boldsymbol{\epsilon}}_v^{\text{a-drv}} + s_v \cdot (\hat{\boldsymbol{\epsilon}}_v^{\text{joint}} - \hat{\boldsymbol{\epsilon}}_v^{\text{a-drv}}), & \hat{\boldsymbol{\epsilon}}^{\text{a-drv}} &= \hat{\boldsymbol{\epsilon}}_\theta(\mathbf{z}_{v,t}, \mathbf{z}_{a,0}^{\varnothing}, \mathbf{c}, \mathbf{b}), \\
    \tilde{\boldsymbol{\epsilon}}_a &= \hat{\boldsymbol{\epsilon}}_a^{\text{v-drv}} + s_a \cdot (\hat{\boldsymbol{\epsilon}}_a^{\text{joint}} - \hat{\boldsymbol{\epsilon}}_a^{\text{v-drv}}), & \hat{\boldsymbol{\epsilon}}^{\text{v-drv}} &= \hat{\boldsymbol{\epsilon}}_\theta(\mathbf{z}_{v,0}^{\varnothing}, \mathbf{z}_{a,t}, \mathbf{c}, \mathbf{b}).
\end{align}
The guidance directions isolate audio-driven visual changes (lip motion, sound-source actions) and vision-driven audio changes (speech rhythm, action sound effects) respectively.
This stage requires 3~forward passes ($\hat{\boldsymbol{\epsilon}}^{\text{joint}}$ shared).
The two stages are complementary: Stage~1 establishes context consistency in early denoising steps, while Stage~2 refines frame-level temporal correspondence in later steps.

\vspace{-1em}
\subsection{Data Pipeline and Training}
\label{sec:data_pipeline}
\vspace{-0.5em}

\begin{figure*}[!t]
    \centering
    \includegraphics[width=\linewidth]{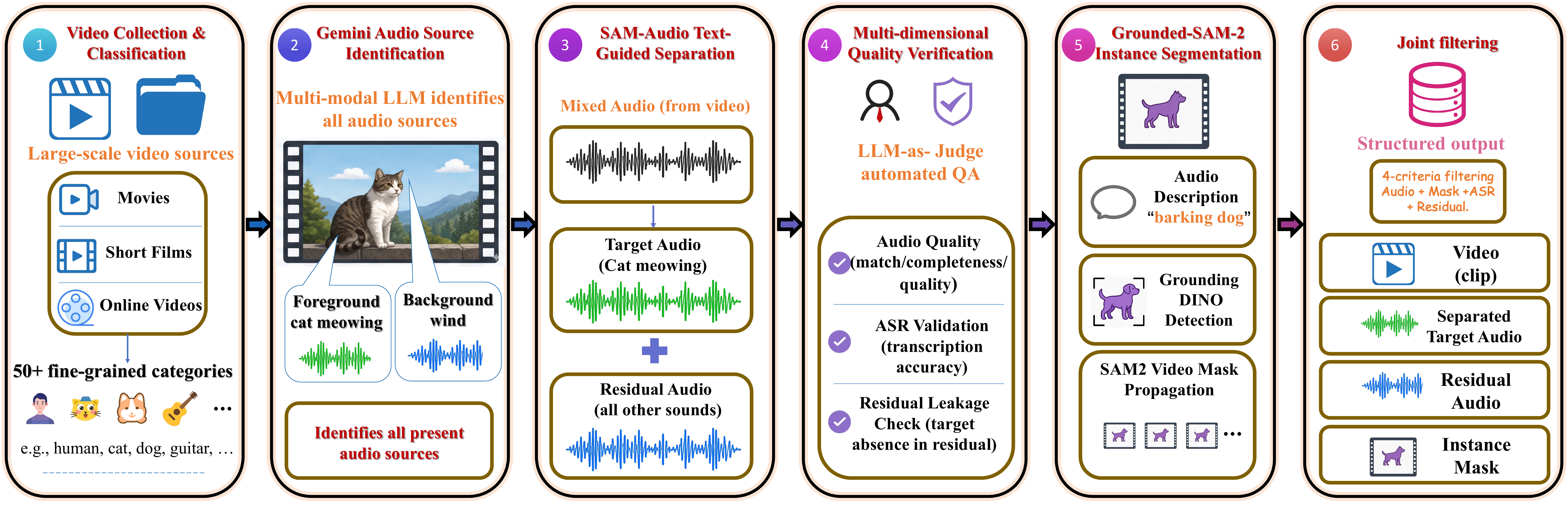}
    \vspace{-15pt}
    \caption{\textbf{Data pipeline.} The pipeline performs six automated stages to convert raw videos into object-level audio-visual editing training pairs without manual annotation.}
    \vspace{-2em}
    \label{fig:datapipe}
\end{figure*}

Synchronized audio-visual editing pairs are scarce for end-to-end training. To address this, we construct a \textbf{scalable data pipeline} (Fig.~\ref{fig:datapipe}) that automatically synthesizes high-quality, object-level editing samples from raw videos without manual annotation.

The pipeline consists of six stages:
(1)~\textbf{Video collection and classification}: Videos are sourced from films, short dramas, and open datasets, filtered by 50+ fine-grained acoustic categories to ensure a single dominant sounding subject.
(2)~\textbf{Multimodal source identification}: Gemini jointly analyzes audio-visual cues to classify sound sources as foreground (target) or background.
(3)~\textbf{Text-guided separation}: SAM-Audio separates the mixture into target and residual (base) audio conditioned on source descriptions.
(4)~\textbf{Multi-dimensional verification}: Gemini assesses separation quality (matching, completeness, leakage, and fidelity); only qualified samples proceed.
(5)~\textbf{Instance segmentation}: Grounding DINO detects the target subject via textual prompts, and SAM2 propagates detections into per-frame masks.
(6)~\textbf{Joint filtering}: Samples are exported upon passing strict criteria for audio quality, mask validity, ASR correctness, and residual cleanliness.

Each pair comprises the original video, editing mask, target/residual audio, reference image, and text description. The original video serves as the reconstruction ground truth, while the mask defines the editing region and base audio provides context for the Context-Aware Module. The resulting dataset contains 400K samples ($\approx$390 hours).

\section{Implementation Details}
\label{sec:suppl_impl}

\subsection{Model Architecture}

The video branch is built on Wan2.2-TI2V-5B: 30-layer DiT blocks, hidden dimension 3072, 24 attention heads, patch size $1{\times}2{\times}2$ (temporal $\times$ spatial). Video VAE compression ratio is temporal $4\times$, spatial $8{\times}8$, with 16 latent channels.
The audio branch adopts the VAE from MMAudio to encode 16\,kHz audio into 2D latent with frequency compression $8\times$, temporal compression $4\times$, and 8 latent channels.

Cross-modal attention adopts local temporal grouping for efficiency: A${\to}$V uses group size 1.25 and window size 3 (covering $\pm$40\,ms perceptual tolerance); V${\to}$A uses group size 0.8 and window size 1. Video Key/Value in V${\to}$A are detached from the computation graph to prevent audio loss from backpropagating into the video stream.

Each audio block contains Visual Context Attention followed by Acoustic Context Attention (both cross-attention layers with zero-initialized output projections; Acoustic Context Attention local window size $w{=}8$).
The condition patch embedding $\mathcal{E}_\text{cond}$ is initialized from $\mathcal{E}_\text{target}$; masked video, reference image, and target latent are concatenated along the temporal dimension into a unified input sequence.

\subsection{Training Configuration}

We train SpongeBob on 240 GPUs (96\,GB each) with a total batch size of 240, learning rate $1{\times}10^{-5}$ with cosine decay. Each training sample consists of 121 frames (approximately 5\,s at 24\,FPS) at 540p resolution. Training runs for 10K steps. The four training modes are sampled with probabilities: Joint Editing 0.4, Audio-driven 0.2, Video-driven 0.2, Context-null 0.2. Condition drop probabilities are 0.1 for both text and base audio (forced to 1.0 in Context-null mode). Mask augmentation applies random dilation up to 20\,px per side with 30\% probability of replacing the precise mask with its bounding box.

\subsection{Inference Configuration}

We use 50 total denoising steps with Flow Matching (linear schedule). Stage 1 (context conflict resolution) runs for steps 1--10 with $s_\text{ctx}{=}5.0$; Stage 2 (temporal synchronization enhancement) runs for steps 11--50 with $s_v{=}5.0$, $s_a{=}5.0$. Negative anchors are: muted audio $\mathbf{z}_{a,0}^{\varnothing}$ (all-zero audio encoded by Audio VAE) and static video $\mathbf{z}_{v,0}^{\varnothing}$ (white image repeated for $T$ frames, encoded by Video VAE). Stage 1 requires 2 forward passes per step and Stage 2 requires 3 (joint prediction shared), totaling $10{\times}2 + 40{\times}3 = 140$ forward passes. Single-sample inference (121 frames at 540p + audio) takes approximately 600\,s on a single H20 GPU.

\section{Experiments}
\vspace{-0.5em}
\subsection{Experimental Setup}
\vspace{-0.5em}

\noindent\textbf{SpongeBob-Bench and evaluation metrics.}
We propose SpongeBob-Bench for systematic evaluation of joint audio-visual editing, comprising 700 test samples across three subsets:
\textbf{Speech-Video} (400 samples) evaluates speaker editing, lip synchronization, and non-target speaker preservation;
\textbf{Sound-Video} (100 samples) evaluates temporal synchronization between object actions and event sounds;
\textbf{Complex Scene} (200 samples) evaluates context consistency when multi-person dialogues, non-target speaker voices, and ambient sounds coexist.

Evaluation metrics span four dimensions.
\textbf{Video quality~\citep{huang2024vbench}}: FVD (Fr\'{e}chet Video Distance), MS (motion smoothness), DD (dynamic degree), and BG (background preservation).
\textbf{Audio quality}: PQ (AudioBox-Aesthetics perceptual quality) and CLAP~\citep{elizalde2023clap} (text-audio semantic alignment).
\textbf{AV synchronization}: Sync-C / Sync-D~\citep{raina2022syncnet} (SyncNet lip-sync) and IB~\citep{girdhar2023imagebind} (ImageBind cross-modal consistency).
\textbf{Context consistency}: Ctx-F1 (based on pyannote~\citep{bredin2020pyannote} speaker detection, jointly penalizing audio conflict and target silence) and G-Score (Gemini~2.5~Pro multimodal holistic score, 1--10).
We additionally evaluate on the external \textbf{AvED-Bench}~\citep{lin2026zero} using its original metric suite (FVD, IS, FC, TC, AC) to verify generalization. Further details on SpongeBob-Bench construction and metric implementation are provided in the supplementary material.

\noindent\textbf{Baselines.}
We compare with four methods representing different paradigms:
(1)~\textbf{AvED}~\citep{lin2026zero}: a zero-shot cross-modal editing method based on pretrained text-to-image and text-to-audio diffusion models;
(2)~\textbf{VACE-Foley}~\citep{jiang2025vace,shan2025hunyuanvideo}: a cascade that first edits video with VACE, then synthesizes audio from scratch with HunyuanVideo-Foley;
(3)~\textbf{VACE-Coherent}~\citep{jiang2025vace,ishii2025coherent}: likewise uses VACE for video editing but employs Coherent to edit audio conditioned on the source audio, preserving source audio structure;
(4)~\textbf{AVI-Edit}~\citep{zheng2025audio}: uses Chatterbox-Turbo (speech TTS) and Stable Audio Open (non-speech SFX) to generate target audio, mixes with residual audio, then drives the AVI-Edit video editing backbone with audio as condition.
All methods use their original pretrained weights and are evaluated on the same test set.

\vspace{-0.5em}
\subsection{Main Results}
\vspace{-0.5em}

\begin{figure*}[!t]
    \centering
    \includegraphics[width=\linewidth]{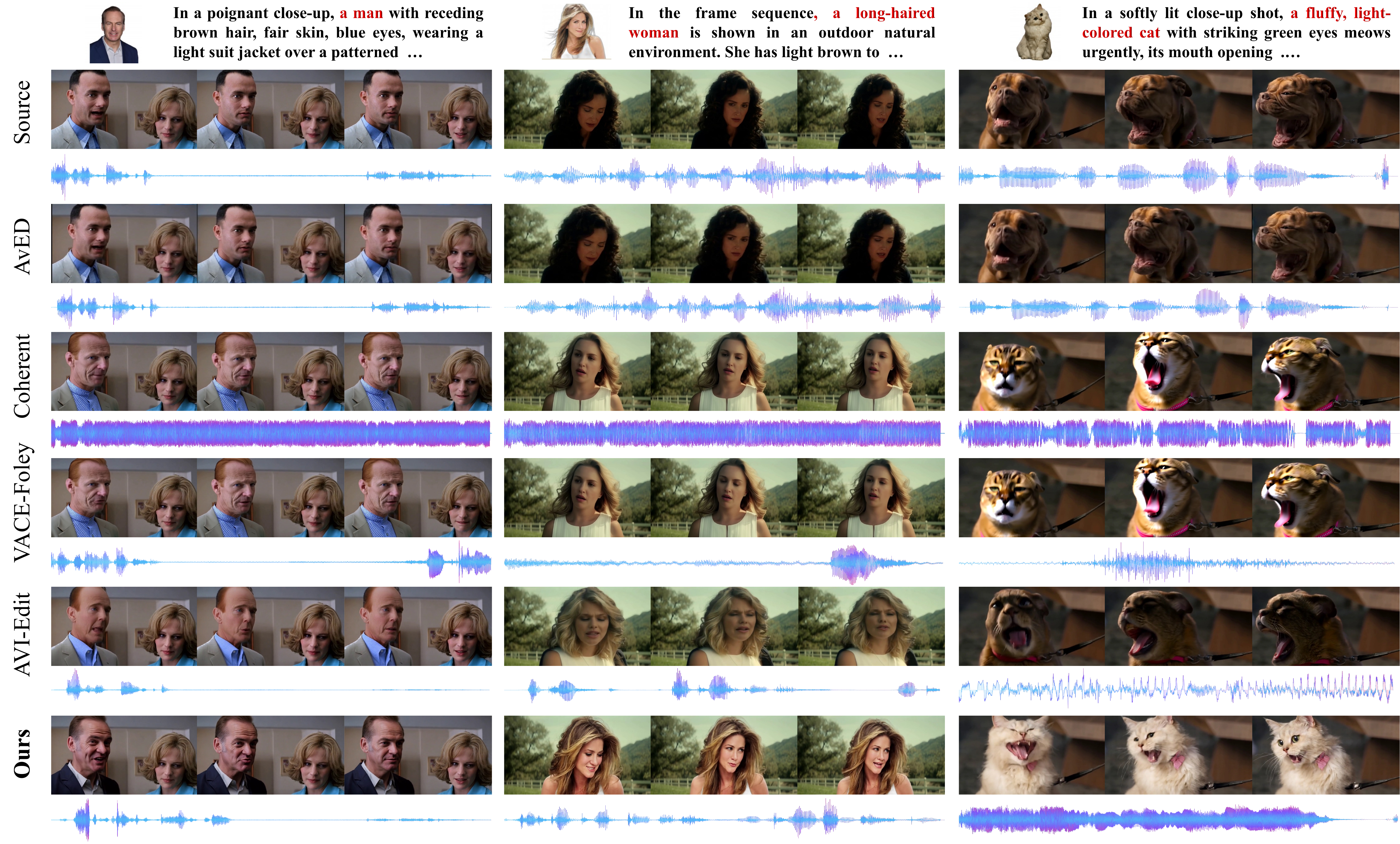}
\caption{\textbf{Multi-scenario qualitative comparison.} SpongeBob achieves faithful visual editing with precisely synchronized audio across all scenarios.}
    \label{fig:qualitative}
\end{figure*}

\noindent\textbf{Qualitative comparison.}
Fig.~\ref{fig:qualitative} compares results on multi-speaker dialogue, single-speaker, and non-speech sound scenarios. SpongeBob achieves synchronized editing across all cases, whereas even AVI-Edit exhibits audio-visual misalignment or context disruption.

\noindent\textbf{SpongeBob-Bench results.}
Table~\ref{tab:main} reports averaged results across the three subsets.
SpongeBob achieves the best performance across all four evaluation dimensions.
In video quality and audio quality, SpongeBob leads comprehensively, demonstrating that end-to-end joint denoising does not sacrifice single-modality quality.
More importantly, the core advantages lie in AV synchronization and context consistency: SpongeBob achieves the highest Sync-C (4.50) and lowest Sync-D (8.73), improving over the strongest baseline AVI-Edit by \textbf{30\%} and \textbf{15.1\%} respectively.
For context, Ctx-F1 improves from 0.72 to 0.81 (+12.5\%), and G-Score from 6.2 to 7.6. Notably, VACE-Foley and VACE-Coherent share identical video metrics (both use VACE-14B, producing identical video output).
VACE-Foley outperforms on audio quality and synchronization (PQ 5.85 vs.\ 5.62, Sync-C 1.85 vs.\ 1.72) thanks to unconstrained generation from video conditions, while VACE-Coherent achieves higher Ctx-F1 (0.68 vs.\ 0.62) due to source audio conditioning that better preserves acoustic structure.
However, both remain far below SpongeBob on synchronization, confirming that source audio conditioning within a cascade cannot compensate for the lack of bidirectional interaction.

\begin{table*}[!t]
\centering
\caption{\textbf{SpongeBob-Bench comprehensive evaluation.} Evaluation across video quality, audio quality, AV synchronization, and context consistency. Best in \textbf{bold}, second best \underline{underlined}.}
\vspace{-0.5em}
\label{tab:main}
\tablefontsize
\setlength{\tabcolsep}{1pt}
\begin{tabular*}{\textwidth}{@{\extracolsep{\fill}}l cccc cc ccc cc@{}}
\toprule
 & \multicolumn{4}{c}{Video Quality} & \multicolumn{2}{c}{Audio Quality} & \multicolumn{3}{c}{AV Sync} & \multicolumn{2}{c}{Context} \\
\cmidrule(lr){2-5} \cmidrule(lr){6-7} \cmidrule(lr){8-10} \cmidrule(lr){11-12}
Method & FVD$\downarrow$ & MS$\uparrow$ & DD$\uparrow$ & BG$\uparrow$ & PQ$\uparrow$ & CLAP$\uparrow$ & Sync-C$\uparrow$ & Sync-D$\downarrow$ & IB$\uparrow$ & Ctx-F1$\uparrow$ & G-Score$\uparrow$ \\
\midrule
AvED          & 548.37 & 0.952 & 0.18 & 0.862 & 4.85 & 0.215 & 1.15 & 12.85 & 0.15 & 0.52 & 3.6 \\
VACE-Foley    & 372.15 & 0.982 & 0.32 & 0.918 & 5.85 & 0.208 & 1.85 & 11.42 & 0.19 & 0.62 & 5.3 \\
VACE+Coh.     & 372.15 & 0.982 & 0.32 & 0.918 & 5.62 & 0.198 & 1.72 & 11.65 & 0.18 & 0.68 & 5.1 \\
AVI-Edit      & \underline{318.56} & \underline{0.985} & \underline{0.35} & \underline{0.932} & \underline{6.12} & \underline{0.225} & \underline{3.45} & \underline{10.28} & \underline{0.21} & \underline{0.72} & \underline{6.2} \\
\rowcolor{oursrow} \textbf{Ours} & \textbf{285.93} & \textbf{0.990} & \textbf{0.36} & \textbf{0.951} & \textbf{6.45} & \textbf{0.238} & \textbf{4.50} & \textbf{8.73} & \textbf{0.24} & \textbf{0.81} & \textbf{7.6} \\
\bottomrule
\end{tabular*}
\end{table*}

\noindent\textbf{AvED-Bench generalization.}
To verify generalization, we evaluate on the external AvED-Bench which focuses on non-speech environmental sound editing.
For fairness and reproducibility, we use each method's original pretrained weights for inference and replace the commercial TTS in AVI-Edit's audio generation module with open-source alternatives.
As shown in Table~\ref{tab:aved}, SpongeBob leads across all metrics: AC improves by 2.9\% (22.15 vs.\ 21.52) and FVD decreases by 5.6\% (338.62 vs.\ 358.47), demonstrating that our framework generalizes beyond speech scenarios.


\vspace{-0.5em}
\subsection{Ablation Study}
\vspace{-0.5em}
\noindent\textbf{Overall Ablations.}
As shown in Table~\ref{tab:ablation_overall}, removing any component causes notable degradation.
Removing Mask routing \& Temporal unification drops Sync-C from 4.50 to 3.18 and BG from 0.951 to 0.915, indicating that without spatial constraints audio-driven visual changes leak into background regions, and without temporal alignment frame-level synchronization degrades.
Removing the Context-Aware Module reduces Ctx-F1 to 0.75 and BG to 0.908, confirming its necessity for resolving context conflicts.
Removing SPTG degrades all metrics: PQ drops to 6.08, BG to 0.935, Sync-C to 3.65, Sync-D rises to 10.12, IB drops to 0.21, and Ctx-F1 to 0.76, validating that SPTG is indispensable for video quality, audio quality, synchronization, and context consistency alike.

\begin{table*}[!t]
\tablefontsize
\setlength{\tabcolsep}{1pt}
\begin{minipage}[t]{0.48\textwidth}
\centering
\caption{\textbf{Overall component ablation.}}
\vspace{-0.5em}
\label{tab:ablation_overall}
\begin{tabular*}{\linewidth}{@{\extracolsep{\fill}}l cccccc@{}}
\toprule
Variant & PQ$\uparrow$ & BG$\uparrow$ & Sync-C$\uparrow$ & Sync-D$\downarrow$ & IB$\uparrow$ & Ctx-F1$\uparrow$ \\
\midrule
w/o M \& T   & 6.28 & 0.915 & 3.18 & 10.52 & 0.20 & 0.78 \\
w/o Ctx Mod  & 6.15 & 0.908 & 4.25 & 9.45 & 0.22 & 0.75 \\
w/o SPTG     & 6.08 & 0.935 & 3.65 & 10.12 & 0.21 & 0.76 \\
\rowcolor{oursrow} \textbf{Full} & \textbf{6.45} & \textbf{0.951} & \textbf{4.50} & \textbf{8.73} & \textbf{0.24} & \textbf{0.81} \\
\bottomrule
\end{tabular*}
\end{minipage}
\hfill
\begin{minipage}[t]{0.48\textwidth}
\centering
\caption{\textbf{Context-aware module ablation.}}
\vspace{-0.5em}
\label{tab:ablation_context}
\begin{tabular*}{\linewidth}{@{\extracolsep{\fill}}l ccccc@{}}
\toprule
Variant & PQ$\uparrow$ & BG$\uparrow$ & Sync-C$\uparrow$ & Sync-D$\downarrow$ & Ctx-F1$\uparrow$ \\
\midrule
No context    & 6.15 & 0.908 & 4.25 & 9.45 & 0.75 \\
Acoustic only & 6.25 & 0.925 & 4.30 & 9.38 & 0.72 \\
Visual only   & 6.18 & 0.932 & 4.35 & 9.32 & 0.80 \\
\rowcolor{oursrow} \textbf{Full (A+V)} & \textbf{6.45} & \textbf{0.951} & \textbf{4.50} & \textbf{8.73} & \textbf{0.81} \\
\bottomrule
\end{tabular*}
\end{minipage}
\end{table*}

\begin{table*}[!t]
\tablefontsize
\setlength{\tabcolsep}{1pt}
\begin{minipage}[t]{0.52\textwidth}
\centering
\vspace{-1em}
\caption{\textbf{SPTG ablation.}}
\vspace{-0.5em}
\label{tab:ablation_sptg}
\begin{tabular*}{\linewidth}{@{\extracolsep{\fill}}l cccccc@{}}
\toprule
Variant & CLAP$\uparrow$ & PQ$\uparrow$ & Sync-C$\uparrow$ & Sync-D$\downarrow$ & BG$\uparrow$ & Ctx-F1$\uparrow$ \\
\midrule
No CFG       & 0.218 & 5.82 & 3.85 & 9.82 & 0.942 & 0.73 \\
Std.\ 2-pass & 0.232 & 6.25 & 3.90 & 9.78 & 0.940 & 0.73 \\
S1 (ctx)     & 0.228 & 6.15 & 3.95 & 9.68 & 0.948 & 0.78 \\
S2 (sync)    & 0.230 & 6.18 & 4.32 & 9.35 & 0.940 & 0.72 \\
\rowcolor{oursrow} \textbf{Full} & \textbf{0.238} & \textbf{6.45} & \textbf{4.50} & \textbf{8.73} & \textbf{0.951} & \textbf{0.81} \\
\bottomrule
\end{tabular*}
\end{minipage}
\hfill
\begin{minipage}[t]{0.46\textwidth}
\centering
\vspace{-1em}
\caption{\textbf{AvED-Bench generalization.}}
\vspace{-0.5em}
\label{tab:aved}
\begin{tabular*}{\linewidth}{@{\extracolsep{\fill}}l ccccc@{}}
\toprule
Method & FVD$\downarrow$ & IS$\uparrow$ & FC$\uparrow$ & TC$\uparrow$ & AC$\uparrow$ \\
\midrule
AvED       & 435.2 & 1.110 & 94.52 & 24.35 & 20.18 \\
VACE-Foley & 418.2 & 1.105 & 95.48 & 25.02 & 21.12 \\
VACE-Coh.  & 418.2 & 1.105 & 95.48 & 25.02 & 21.42 \\
AVI-Edit   & \underline{358.5} & \underline{1.120} & \underline{95.68} & \underline{25.18} & \underline{21.52} \\
\rowcolor{oursrow} \textbf{Ours} & \textbf{338.6} & \textbf{1.130} & \textbf{96.05} & \textbf{25.45} & \textbf{22.15} \\
\bottomrule
\end{tabular*}
\end{minipage}
\end{table*}

\begin{figure}[!t]
    \centering
    \vspace{-1em}
    \includegraphics[width=\linewidth]{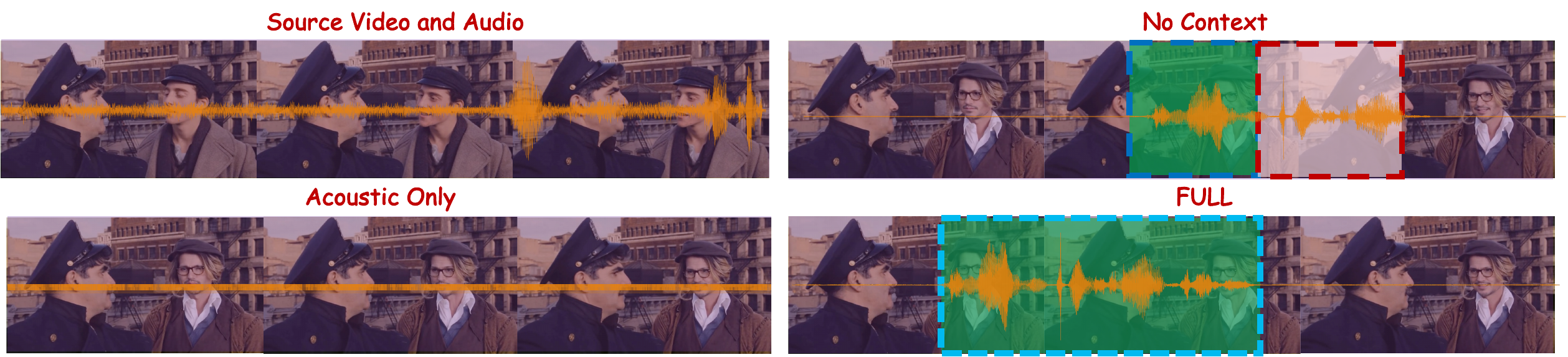}
\caption{\textbf{Context conflict visualization.} Red indicates overlap with non-target speech; green indicates correct generation. The full module precisely avoids non-target speech while generating normally during ambient sounds.}
    \vspace{-2em}
    \label{fig:context_conflict}
\end{figure}

\noindent\textbf{Necessity of Context-Aware Module.}
As shown in Table~\ref{tab:ablation_context}, without any context module (No context), Ctx-F1 is 0.75 due to heavy temporal overlap between target and non-target speech.
With Visual Context Attention alone, the model leverages lip movements in the masked video to localize non-target speech intervals, achieving selective avoidance and raising Ctx-F1 to 0.80.
However, with Acoustic Context Attention alone, although PQ and BG improve, Ctx-F1 drops to 0.72 (below No context).
As illustrated in Fig.~\ref{fig:context_conflict}, since base audio after source separation remains a mixed signal, the model cannot distinguish non-target speech from ambient sound and thus conservatively avoids all acoustically active intervals, causing excessive target silence.
The full configuration combines both: Visual Context Attention provides cross-modal disambiguation to localize non-target speech, while Acoustic Context Attention provides the original acoustic environment as reference, yielding the best Ctx-F1 of 0.81.


\noindent\textbf{Efficacy of SPTG.}
Table~\ref{tab:ablation_sptg} compares five inference guidance strategies on the same trained model.
Without any guidance (No CFG), single-modality quality is low (PQ~5.82, CLAP~0.218, BG~0.942).
Standard 2-pass CFG raises PQ and CLAP but barely improves synchronization or context (Sync-C 3.85$\to$3.90, Ctx-F1 0.73$\to$0.73), confirming that standard CFG primarily enhances single-modality adherence with limited cross-modal benefit.
Stage~1 alone (context CFG) raises Ctx-F1 to 0.78 and BG to 0.948.
Stage~2 alone (sync CFG) raises Sync-C to 4.32 and lowers Sync-D to 9.35, but Ctx-F1 slightly decreases (0.72).
Full SPTG combining both stages achieves the best across all metrics (Ctx-F1~0.81, Sync-C~4.50), demonstrating complementary enhancement.


\section{Conclusion}

We present \textbf{SpongeBob}, an end-to-end audio-visual joint editing framework based on bidirectional cross-modal interaction that simultaneously edits visual content and synthesizes synchronized audio.
Cascaded paradigms suffer from desynchronization, background audio loss, and spatially unaware editing due to the lack of cross-modal information exchange between their independent stages.
SpongeBob addresses these issues through three core designs:
the Sync-Aware Editing Mechanism aligns visual motion and sound events from interaction, temporal, and spatial dimensions;
the Context-Aware Module perceives unedited audio-visual context to prevent conflicts with preserved content;
and SPTG protects cross-modal synchronization and context consistency at both training and inference stages.
Together with \textbf{SpongeBob-Bench}, these contributions demonstrate that joint audio-visual editing requires carefully designed cross-modal information flow across spatial, temporal, and acoustic dimensions, rather than simply combining two separate models.



\label{references}
\addcontentsline{toc}{section}{References}
\bibliographystyle{assets/plainnat}
\bibliography{paper}

\appendix
\newpage
\appendix



\section{SpongeBob-Bench Construction and Metrics}
\label{sec:suppl_bench}

\subsection{Construction Pipeline}

SpongeBob-Bench is constructed through four steps: (1) candidate samples are collected from independent video sources with no overlap with the training set; (2) the same automated quality verification as training data is applied (audio separation quality, mask quality, ASR validity, residual cleanliness); (3) human review confirms ground truth accuracy, mask precision, and text description correctness; (4) samples are assigned to three subsets based on scene characteristics, ensuring intra-subset diversity. The final benchmark contains 700 test samples with strict non-overlap from the training set.

\subsection{Scene Partitioning}

\begin{table}[ht]
\centering
\tablefontsize
\begin{tabular}{@{}lrl@{}}
\toprule
Subset & Samples & Core Evaluation Target \\
\midrule
Speech-Video & 400 & Lip sync, speaker editing, non-target preservation \\
Sound-Video & 100 & Action-sound temporal alignment \\
Complex Scene & 200 & Multi-source context consistency \\
\bottomrule
\end{tabular}
\end{table}

\subsection{Metric Computation}

\textbf{Video quality.}
FVD measures Fr\'{e}chet distance of I3D features.
MS computes mean cosine similarity of DINOv2 features between adjacent frames.
DD measures mean RAFT optical flow magnitude within the editing region.
BG computes cosine similarity of DINOv2 features outside the mask before and after editing.

\textbf{Audio quality.}
PQ uses AudioBox-Aesthetics for perceptual quality scoring (1--10).
CLAP computes cosine similarity between CLAP audio and text embeddings.

\textbf{AV synchronization.}
Sync-C and Sync-D measure SyncNet lip-sync confidence and distance respectively (only for subsets containing speakers).
IB computes mean per-frame cosine similarity of ImageBind audio-visual embeddings.

\textbf{Context consistency.}
Ctx-F1 is based on pyannote speaker detection: Precision $= 1 -$ conflict rate (overlap between generated target audio and non-target speakers); Recall $=$ target audio coverage rate; F1 combines both, simultaneously penalizing conflicts and excessive silence.
G-Score uses Gemini 2.5 Pro as a multimodal evaluator providing holistic scores (1--10), averaged over 3 evaluations per sample.

\section{Per-Scene Detailed Results}
\label{sec:suppl_perscene}

Table~\ref{tab:main} in the main paper reports weighted averages across the three subsets (weights 400:100:200). Sync-C, Sync-D, and Ctx-F1 are computed only on subsets containing speakers (Speech 400 + Complex 200 = 600). Below we provide per-subset breakdowns.

\begin{table*}[t]
\centering
\caption{\textbf{Speech-Video subset (400 samples).} SpongeBob achieves +29\% Sync-C over AVI-Edit, benefiting from bidirectional cross-modal attention that enables lip generation to perceive audio phoneme rhythms in real time.}
\label{tab:speech_video}
\tablefontsize
\begin{tabular}{@{}l ccccccccccc@{}}
\toprule
Method & FVD$\downarrow$ & MS$\uparrow$ & DD$\uparrow$ & BG$\uparrow$ & PQ$\uparrow$ & CLAP$\uparrow$ & Sync-C$\uparrow$ & Sync-D$\downarrow$ & IB$\uparrow$ & Ctx-F1$\uparrow$ & G$\uparrow$ \\
\midrule
AvED       & 545.0 & 0.953 & 0.18 & 0.865 & 4.90 & 0.215 & 1.28 & 12.55 & 0.16 & 0.55 & 3.6 \\
VACE-Foley & 365.0 & 0.983 & 0.32 & 0.920 & 5.88 & 0.208 & 2.05 & 11.15 & 0.20 & 0.65 & 5.4 \\
VACE+Coh.  & 365.0 & 0.983 & 0.32 & 0.920 & 5.65 & 0.198 & 1.92 & 11.35 & 0.19 & 0.72 & 5.2 \\
AVI-Edit   & 310.0 & 0.986 & 0.35 & 0.935 & 6.15 & 0.225 & 3.85 & 9.85  & 0.22 & 0.75 & 6.3 \\
\rowcolor{oursrow} \textbf{Ours} & \textbf{280.5} & \textbf{0.991} & \textbf{0.36} & \textbf{0.952} & \textbf{6.48} & \textbf{0.240} & \textbf{4.95} & \textbf{8.25} & \textbf{0.24} & \textbf{0.84} & \textbf{7.7} \\
\bottomrule
\end{tabular}
\end{table*}

\begin{table*}[t]
\centering
\caption{\textbf{Sound-Video subset (100 samples).} No lip sync is involved; IB Score serves as the primary synchronization metric. SpongeBob achieves IB 0.28 vs.\ AVI-Edit 0.24 (+17\%).}
\label{tab:sound_video}
\tablefontsize
\begin{tabular}{@{}l cccccccc@{}}
\toprule
Method & FVD$\downarrow$ & MS$\uparrow$ & DD$\uparrow$ & BG$\uparrow$ & PQ$\uparrow$ & CLAP$\uparrow$ & IB$\uparrow$ & G$\uparrow$ \\
\midrule
AvED       & 520.0 & 0.958 & 0.24 & 0.878 & 5.15 & 0.238 & 0.19 & 4.0 \\
VACE-Foley & 348.0 & 0.986 & 0.38 & 0.928 & 6.05 & 0.228 & 0.22 & 5.8 \\
VACE+Coh.  & 348.0 & 0.986 & 0.38 & 0.928 & 5.82 & 0.218 & 0.21 & 5.5 \\
AVI-Edit   & 295.0 & 0.988 & 0.40 & 0.942 & 6.32 & 0.242 & 0.24 & 6.6 \\
\rowcolor{oursrow} \textbf{Ours} & \textbf{268.2} & \textbf{0.993} & \textbf{0.42} & \textbf{0.958} & \textbf{6.62} & \textbf{0.258} & \textbf{0.28} & \textbf{8.0} \\
\bottomrule
\end{tabular}
\end{table*}

\begin{table*}[t]
\centering
\caption{\textbf{Complex Scene subset (200 samples).} The most challenging subset with all metrics lower than other subsets. SpongeBob improves Ctx-F1 by +14\% over AVI-Edit, a larger margin than in Speech-Video (+12\%), demonstrating greater contribution of the Context-Aware Module in complex multi-source scenarios.}
\label{tab:complex_scene}
\tablefontsize
\begin{tabular}{@{}l ccccccccccc@{}}
\toprule
Method & FVD$\downarrow$ & MS$\uparrow$ & DD$\uparrow$ & BG$\uparrow$ & PQ$\uparrow$ & CLAP$\uparrow$ & Sync-C$\uparrow$ & Sync-D$\downarrow$ & IB$\uparrow$ & Ctx-F1$\uparrow$ & G$\uparrow$ \\
\midrule
AvED       & 569.3 & 0.947 & 0.15 & 0.848 & 4.60 & 0.204 & 0.89 & 13.45 & 0.11 & 0.46 & 3.4 \\
VACE-Foley & 398.5 & 0.978 & 0.29 & 0.909 & 5.69 & 0.198 & 1.45 & 11.96 & 0.16 & 0.56 & 4.85 \\
VACE+Coh.  & 398.5 & 0.978 & 0.29 & 0.909 & 5.46 & 0.188 & 1.32 & 12.25 & 0.15 & 0.60 & 4.70 \\
AVI-Edit   & 347.5 & 0.982 & 0.32 & 0.921 & 5.96 & 0.216 & 2.65 & 11.14 & 0.18 & 0.66 & 5.80 \\
\rowcolor{oursrow} \textbf{Ours} & \textbf{305.7} & \textbf{0.987} & \textbf{0.33} & \textbf{0.945} & \textbf{6.30} & \textbf{0.224} & \textbf{3.60} & \textbf{9.69} & \textbf{0.22} & \textbf{0.75} & \textbf{7.2} \\
\bottomrule
\end{tabular}
\end{table*}

\section{User Study}
\label{sec:suppl_user}

We conduct a user study to validate SpongeBob's perceptual advantages as a complement to automatic metrics. We randomly sample 30 test cases from SpongeBob-Bench (15 Speech-Video, 5 Sound-Video, 10 Complex Scene) and compare SpongeBob with all four baselines. Twenty evaluators with audio-visual professional backgrounds independently view all results in randomized anonymous order and rate each on a 1--5 scale across four dimensions: \textbf{AV-Sync} (temporal alignment naturalness), \textbf{Audio-Q} (clarity and realism), \textbf{Context} (whether new sounds conflict with preserved background audio and non-target speakers), and \textbf{Overall} (holistic editing quality).

\begin{table}[ht]
\centering
\caption{\textbf{User study (MOS, 1--5).} SpongeBob significantly outperforms all baselines ($p < 0.01$, paired $t$-test).}
\label{tab:user_study}
\tablefontsize
\begin{tabular}{@{}lcccc@{}}
\toprule
Method & AV-Sync$\uparrow$ & Audio-Q$\uparrow$ & Context$\uparrow$ & Overall$\uparrow$ \\
\midrule
AvED       & 2.12 & 2.45 & 2.28 & 2.18 \\
VACE-Foley & 2.98 & 3.25 & 2.68 & 2.88 \\
VACE+Coh.  & 2.85 & 3.18 & 3.05 & 2.92 \\
AVI-Edit   & 3.42 & 3.65 & 3.28 & 3.38 \\
\rowcolor{oursrow} \textbf{Ours} & \textbf{4.28} & \textbf{4.15} & \textbf{4.32} & \textbf{4.25} \\
\bottomrule
\end{tabular}
\end{table}

SpongeBob significantly outperforms all baselines across all dimensions ($p < 0.01$, paired $t$-test). Compared with the strongest baseline AVI-Edit, the advantage is most pronounced on AV-Sync (+0.86) and Context (+1.04), consistent with automatic metric trends. Notably, VACE-Foley scores higher than VACE+Coh.\ on AV-Sync and Audio-Q (consistent with its higher Sync-C and PQ in automatic metrics), but lower on Context (2.68 vs.\ 3.05) since generating audio from scratch without source audio conditioning fails to respect the existing acoustic scene.

\section{Data Pipeline Details}
\label{sec:suppl_data}

The main paper (\S3.4) outlines the six-stage pipeline. Here we supplement key technical details.

\paragraph{Fine-grained acoustic category taxonomy.}
We design 50+ fine-grained categories covering animals (cat/dog/bird/frog breeds), instruments (string/wind/percussion/ethnic), and human speech. Each category precisely describes the sounding subject type and acoustic characteristics, used for candidate video retrieval and driving subsequent separation.

\paragraph{Quality verification.}
Gemini executes all quality assessments in an ``LLM-as-Judge'' paradigm across five dimensions: match score ($\geq$5, semantic consistency), completeness ($\geq$5, full extraction), quality ($\geq$5, no artifacts), leakage score ($\geq$ threshold, no target leakage in residual), and ASR validity (speech scenes only). Only samples passing all criteria enter the final dataset.

\paragraph{Instance segmentation.}
Source description text $\to$ Grounding DINO first-frame detection (box threshold 0.4, text threshold 0.3) $\to$ SAM2 first-frame segmentation $\to$ SAM2 full-video mask propagation $\to$ Gemini mask quality verification.

\paragraph{Dataset statistics.}
The final dataset contains approximately 400K samples totaling 390 hours at 24 FPS. Speech samples account for approximately 60\% and animal/instrument/environmental samples for 40\%, spanning 50+ fine-grained categories.

\section{Limitations}
\label{sec:suppl_limitations}

(1)~\textbf{Cross-category generalization boundary.} Training is based on mask-and-reconstruct self-supervision where the model only sees same-category reconstruction. Cross-category editing (e.g., dog$\to$cat) relies on compositional generalization of text conditions and reference images; quality may degrade when visual appearance differs substantially.
(2)~\textbf{Inference overhead.} The two-stage SPTG guidance requires 140 total forward passes (vs.\ 100 for standard CFG), increasing inference time by approximately 40\%, which may bottleneck real-time or interactive applications.
(3)~\textbf{Long video limitation.} Training clips are limited to 121 frames (approximately 5\,s); longer videos require segmented processing, potentially introducing audio-visual discontinuities at segment boundaries.

\section{Ethics Statement and Broader Impact}
\label{sec:suppl_ethics}

\paragraph{Positive impact.}
Improved efficiency in film post-production dubbing and sound replacement; accessible content creation; cross-lingual localization of educational videos; creative content production tools.

\paragraph{Potential risks and mitigation.}
Audio-visual editing technology carries risks of deepfake generation, unauthorized content tampering, and erosion of public trust in video authenticity. We mitigate these through: (a) imperceptible digital watermarks in generated content for provenance verification; (b) co-development and open-sourcing of detection models; (c) usage terms prohibiting identity forgery and disinformation; (d) tiered access control for high-risk functionalities.

\paragraph{Data ethics.}
Training data is sourced from publicly available videos with no personal privacy information. Human faces will be anonymized upon dataset release. The pipeline is fully automated, requiring no human workers to process sensitive content.

\end{document}